\documentclass[10pt,twocolumn,letterpaper]{article}

\usepackage{cvpr}              
\usepackage[ruled,vlined]{algorithm2e}
\usepackage{url}
\usepackage{multirow}
\usepackage[dvipsnames]{xcolor}

\definecolor{cvprblue}{rgb}{0.21,0.49,0.74}
\definecolor{azure}{rgb}{0.0, 0.5, 1.0}
\definecolor{amaranth}{rgb}{0.9, 0.17, 0.31}
\definecolor{Gray}{gray}{0.9}
\definecolor{green}{rgb}{0.55, 0.71, 0.0}
\definecolor{amber}{rgb}{1.0, 0.49, 0.0}
\definecolor{byzantine}{rgb}{0.74, 0.2, 0.64}
\definecolor{forestGreen}{rgb}{0.0, 0.54, 0.0}
\definecolor{blue}{rgb}{0.43, 0.71, 0.88}
\definecolor{pink}{rgb}{0.85, 0.44, 0.58}
\definecolor{yellow}{rgb}{0.96, 0.83, 0.26}

\usepackage[pagebackref,breaklinks,colorlinks,citecolor=cvprblue]{hyperref}

\def\ie{\textit{i.e.}}
\def\eg{\textit{e.g.}}
\def\etc{\textit{etc.}}

\newcommand{\mname}[1]{RoHM}
\newcommand{\myparagraph}[1]{\noindent\textbf{#1}}

\def\confName{CVPR}
\def\confYear{2024}

\title{RoHM: Robust Human Motion Reconstruction via Diffusion}

\newcommand*{\affaddr}[1]{#1} 
\newcommand*{\affmark}[1][*]{\textsuperscript{#1}}
\newcommand*{\email}[1]{\small{\texttt{#1}}}

\author{
Siwei Zhang\affmark[1,2]$^*$ \quad
Bharat Lal Bhatnagar\affmark[2] \quad
Yuanlu Xu\affmark[2] \quad
Alexander Winkler\affmark[2] \quad
Petr Kadlecek\affmark[2] \quad \\
Siyu Tang\affmark[1] \quad
Federica Bogo\affmark[2] \quad \\
\affaddr{\affmark[1]ETH Z\"urich} \quad \affaddr{\affmark[2]Meta Reality Labs Research}  \\
\email{\{siwei.zhang, siyu.tang\}@inf.ethz.ch} \\ \email{\{bharatbhatnagar, yuanluxu, winklera, petr.kadlecek, fbogo\}@meta.com}
}

\begin{document}

\twocolumn[{
\maketitle
\begin{center}
    \vspace{-6mm}
    \captionsetup{type=figure}
    \includegraphics[width=\textwidth]{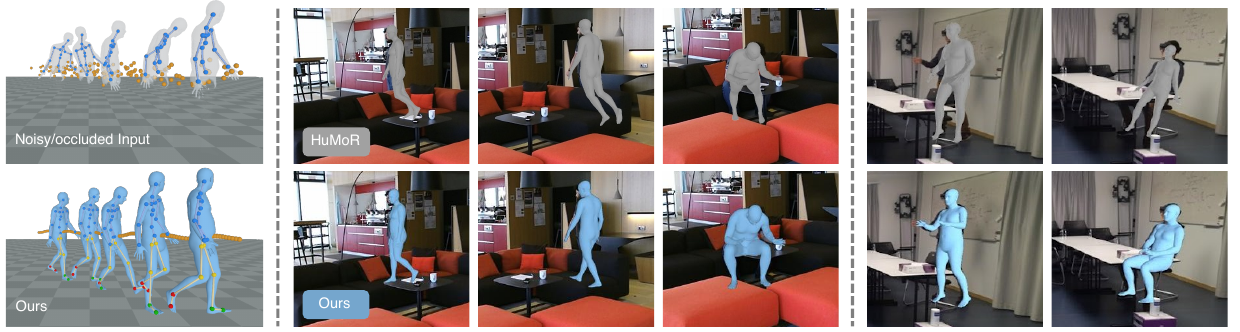}
    \vspace{-7mm}
    \captionof{figure}{Our method robustly reconstructs smooth and complete 3D human motion from different inputs, such as incomplete and noisy motion estimates (left), RGB-D (middle) and RGB (right) monocular videos. We learn diffusion-based models to denoise and infill both \textcolor{orange}{root trajectory} in global space and local motion in body-root space for \textcolor{blue}{visible} and \textcolor{yellow}{occluded} joints, predicting whether feet are \textcolor{forestGreen}{in contact} or \textcolor{red}{not} with the ground for improved physical plausibility. Compared with baselines such as HuMoR~\cite{rempe2021humor}, our method reconstructs more plausible motions that faithfully match image evidence, especially under heavy occlusions.}
    \label{fig:teaser}
\end{center}
}]

\def\thefootnote{*}\footnotetext{The work was done during an internship at Meta.}

\maketitle
\begin{abstract}
\vspace{-3mm}

We propose \mname{}, an approach for robust 3D human motion reconstruction from monocular RGB(-D) videos in the presence of noise and occlusions.
Most previous approaches either train neural networks to directly regress motion in 3D or learn data-driven motion priors and combine them with optimization at test time. The former do not recover globally coherent motion and fail under occlusions; the latter are time-consuming, prone to local minima, and require manual tuning. To overcome these shortcomings, we exploit the iterative, denoising nature of diffusion models.
\mname{} is a novel diffusion-based motion model that, conditioned on noisy and occluded input data, reconstructs complete, plausible motions in consistent global coordinates.
Given the complexity of the problem -- requiring one to address different tasks (denoising and infilling) in different solution spaces (local and global motion) -- we decompose it into two sub-tasks and learn two models, one for global trajectory and one for local motion. To capture the correlations between the two, we then introduce a novel conditioning module, combining it with an iterative inference scheme.
We apply \mname{} to a variety of tasks -- from motion reconstruction and denoising to spatial and temporal infilling. Extensive experiments on three popular datasets show that our method outperforms state-of-the-art approaches qualitatively and quantitatively, while being faster at test time. 
The code is available at \url{https://sanweiliti.github.io/ROHM/ROHM.html}.

\end{abstract}    
\vspace{-6mm}
\section{Introduction}
\label{sec:intro}

In this paper, we tackle the problem of 3D human motion reconstruction from monocular RGB(-D) videos in real scenarios -- \ie, in the presence of noise and occlusions.
Reconstructing 3D human motions is crucial for many applications, ranging from augmented and virtual reality to robotics.
Many methods in the literature tackle the problem by training deep neural networks to directly regress 3D body pose and shape from monocular input~\cite{kocabas2020vibe,kanazawa2019learning,nam2023cyclic,choi2021beyond,goel2023humans}.
However, these approaches commonly suffer from two major shortcomings: 1) they estimate only \emph{local} motion, representing pose in body-root relative coordinates without plausible \emph{global} motion, in world coordinates consistent over time; 2) they lack robustness when the body undergoes occlusions, in either the spatial or temporal dimension.

In such scenarios, optimization-based methods~\cite{rempe2021humor,phasemp,zhang2021learning} have shown better performance. 
For instance, HuMoR~\cite{rempe2021humor} and PhaseMP~\cite{phasemp} explicitly address scenarios with noisy and incomplete input by combining data-driven motion priors with application-agnostic optimizers such as L-BFGS~\cite{nocedal}.
Still, these methods may fail under heavy occlusions (Fig.~\ref{fig:teaser}). 
Furthermore, test-time optimization is time-consuming, prone to local minima, and requires significant manual tuning~\cite{Choutas:ECCV:2022,song2020lgd}.

To overcome these limitations, we propose to leverage the iterative, generative nature of diffusion models. 
Diffusion models were initially proposed for 2D generation tasks~\cite{dhariwal2021diffusion, song2020denoising, ho2020denoising, ho2022classifierfree}; recently, they achieved compelling results in 3D human motion generation from input such as text and action labels~\cite{shafir2023priorMDM,tevet2023human,yuan2023physdiff,karunratanakul2023guided}, music~\cite{edge2023}, and sparse (noise-free) keypoints~\cite{du2023agrol,castillo2023bodiffusion, shafir2023priorMDM}. In particular, these models proved effective in learning and modelling long-term motion dependencies over time~\cite{tevet2023human}, which go beyond the single-frame conditioning considered in~\cite{rempe2021humor,phasemp}. Furthermore, their iterative denoising process poses them as a data-driven alternative to the iterative minimization of optimization-based techniques.
However, so far diffusion-based approaches have mostly focused on \emph{synthesizing} motion from user input, rather than \emph{reconstructing} motion from monocular videos exhibiting noise and occlusions -- where reconstructions need to match image evidence whenever available.
Here, we explore how to leverage diffusion models in such reconstruction scenarios. 

Given a monocular video, we obtain initial, per-frame 3D pose estimates using off-the-shelf regressors~\cite{li2022cliff,feng2021collaborative, sarandi2021metrabs} and/or per-frame optimization (similar to ~\cite{rempe2021humor,phasemp, zhang2021learning}). These estimates are inaccurate and incomplete, with implausible motions. From them, our goal is to reconstruct a smooth and complete 3D motion. This is a complex task, requiring us to address different problems (motion denoising and infilling) in different solution spaces (local and global motion). 
We observe that previously proposed diffusion-based motion models~\cite{shafir2023priorMDM, tevet2023human} do not work well in this scenario: they expect noise-free, user-provided keypoints as input, model global and local motion together, and cannot ensure alignment between reconstruction and image evidence.
Inspired by~\cite{yuan2022glamr,sun2023trace}, we therefore propose to decompose the problem. We leverage motions from the AMASS dataset~\cite{AMASS} to learn two diffusion models conditioned on noisy input, one for global trajectory reconstruction and one for local motion prediction -- showing the benefits of decoupling the two spaces. Still, this formulation ignores the correlations between global and local motion space.
While~\cite{yuan2022glamr} addresses this by predicting trajectory based only on infilled local motion, in our scenario we require the estimated trajectory to remain faithful to the input. Drawing inspiration from~\cite{zhang2023adding}, we therefore propose a flexible conditioning strategy for trajectory reconstruction, exploiting both input data and denoised local motion. We show that this module, combined with an iterative scheme at inference time, improves both local and global motion quality.
Finally, to further encourage physically plausible motions that match image evidence, we guide sampling at test time with physics-based and image-based scores.

In summary, our contributions are:
1) \textbf{\mname{}}, a novel diffusion-based approach for \textbf{Ro}bust \textbf{H}uman \textbf{M}otion reconstruction in consistent global coordinates from monocular sequences with noise and occlusions;
2) a flexible conditioning strategy to capture inter-dependencies between root trajectory and local pose; 
3) various applications enabled by \mname{}, including motion reconstruction, denoising, spatial and temporal infilling.
Extensive experiments on three widely used datasets show that RoHM achieves superior accuracy and realism compared to state-of-the-art optimization-based methods, while being 30 times faster at inference time.

\section{Related Work}

\myparagraph{Regression-based methods}. 
Many approaches in the literature focus on 3D human shape and pose reconstruction from a single image~\cite{kanazawa2018end, kolotouros2019cmr, lassner2017unite, xu2019denserac, zhang2021body,li2021hybrik, lin2021end-to-end, Kocabas_SPEC_2021, cho2022cross, li2022cliff, kolotouros2019convolutional, fang2021reconstructing, kolotouros2019learning, song2020lgd, wang20233d, feng2021collaborative, lin2023one}, recently also considering robustness to occlusions~\cite{zhang20233d, li2023jotr, khirodkar2022occluded, liu2022explicit, Kocabas_PARE_2021, zhang2023probabilistic, Rockwell2020}.
Dealing with occlusions is even more challenging for monocular human motion reconstruction, requiring one to model plausible dynamics over time for non-visible body parts. 
Many approaches train neural networks to predict 3D motion from RGB videos and are not easily adaptable to different input modalities~\cite{cheng19occlusion, kanazawa2019learning, kocabas2020vibe, choi2021beyond, goel2023humans, sun2019human, luo20203d, zanfir2020weakly, you2023co, nam2023cyclic, foo2023distribution, wei2022capturing, rajasegaran2022tracking, pavlakos2022human}.
Some methods introduce adversarial priors~\cite{kanazawa2019learning, kocabas2020vibe}, leverage multi-view cues~\cite{pavlakos2022human} or learn denoising models~\cite{nam2023cyclic} to achieve robustness. 
Still, most of them estimate only local motion in the camera frame without recovering global trajectory, thus suffering from jitter and motion artifacts. \cite{yuan2022glamr,li2022dnd,sun2023trace} estimate plausible global trajectories from per-frame local features, but are not robust to occlusions~\cite{ye2023slahmr}.
In contrast, our method robustly recovers both global and local motion and can be applied to different inputs.

\myparagraph{Optimization-based methods}.
These methods typically fit a parametric body model~\cite{loper2015smpl} to observations (such as body keypoints, depth, silhouettes \etc) by iteratively minimizing an objective function~\cite{Bogo:ECCV:2016, pavlakos2019expressive, rempe2021humor, phasemp, zhang2021learning, ye2023slahmr}. To regularize motion, such function contains one or more temporal priors.
Some approaches define hand-crafted priors encouraging motion smoothness (\eg, on body joint velocity or acceleration)~\cite{arnab2019exploiting, mehta2017vnect}, or constrain motion in a low-dimensional space~\cite{huang2017towards}; however, they may produce over-smooth motions and foot skating~\cite{rempe2021humor}.
Lately, the availability of large-scale motion capture datasets such as AMASS~\cite{AMASS} made it possible to learn powerful motion models from data~\cite{rempe2021humor, phasemp, zhang2021learning}.
LEMO~\cite{zhang2021learning} learns two separate, fully deterministic priors for motion smoothness and infilling.
HuMor~\cite{rempe2021humor} and PhaseMP~\cite{phasemp} propose generative priors, modelling the distribution of state transitions between frames via Conditional Variational Autoencoders (CVAEs)~\cite{cai2021unified,motionvae}. Learning transitions only between pair of frames, these methods struggle when occlusions span long time intervals (see Sec.~\ref{sec:experiment}).
Optimization-based methods tend to match input data more closely than regression methods, thanks to their iterative minimization; but they are in general slower, prone to local minima, and require significant manual tuning~\cite{Choutas:ECCV:2022,song2020lgd}.
In contrast, we propose to leverage the iterative nature of diffusion models.

\myparagraph{Human motion models.} A variety of approaches has been proposed for motion tracking and synthesis, including mixtures-of-Gaussians~\cite{howe2000}, Gaussian processes~\cite{urtasun2006}, pose embeddings~\cite{ormoneit2000learning,pavlovic2001,taylor2007,urtasun2006cviu}, VAEs~\cite{zhang2022wanderings,motionvae}, 2D CNN~\cite{kaufmann2020convolutional}, and normalizing flows~\cite{moglow2020}. These methods may not generalize well to unseen motions~\cite{rempe2021humor} and body-scene interactions.
Physics-based methods~\cite{gartner2022trajectory, huang2022neural, gartner2022differentiable, luo20203d, Luo2022EmbodiedSH, rempe2020contact, yuan2021simpoe, shimada2020physcap, xie2021physics} address these challenges by enforcing physics laws via simulation. However, simulators are computationally expensive, non-differentiable, and may introduce discrepancies between predicted motions and observed input.

\myparagraph{Diffusion models for human motion.}
Diffusion models have demonstrated compelling results for human motion synthesis conditioned on input such as text and action labels~\cite{tevet2023human, yuan2023physdiff, wang2023fg, zhang2023remodiffuse, chen2023executing, karunratanakul2023guided, jiang2024motiongpt, karunratanakul2024dno}, music~\cite{edge2023}, environment~\cite{huang2023diffusion,rempe2023trace}, and (noise-free) 3D joints~\cite{shafir2023priorMDM}.
Given their ability to model long-range spatio-temporal relationships, they have been applied to motion forecasting~\cite{barquero2023belfusion, tanke2023social} and infilling~\cite{tevet2023human,shafir2023priorMDM}.
~\cite{du2023agrol,castillo2023bodiffusion} use diffusion to synthesize lower body motion given head and wrist positions and rotations.
These methods do not focus on scenarios with noisy input or body occlusion varying over time.
Recently, diffusion has been leveraged for 3D body pose estimation ~\cite{foo2023distribution,gong2022diffpose}, even under severe body truncations~\cite{zhang2023probabilistic}, but without considering the temporal dimension.
In contrast, we leverage diffusion models to achieve robustness against varying ranges of occlusion and noise over long temporal sequences.

\section{Preliminaries}
\label{sec:preliminaries}
\myparagraph{SMPL-X ~\cite{pavlakos2019expressive}} is a parametric model that represents the human body as a function $\mathcal{M}(\boldsymbol{\gamma}, \boldsymbol{\Phi}, \boldsymbol{\theta}, \boldsymbol{\beta}, \boldsymbol{{\theta}}_h, \boldsymbol{\phi})$, parameterized by global translation $\boldsymbol{\gamma}$, global orientation $\boldsymbol{\Phi}$, body pose $\boldsymbol{\theta}$, body shape $\boldsymbol{\beta}$, hand pose $\boldsymbol{{\theta}_h}$ and facial expression $\boldsymbol{\phi}$. The function returns a triangle mesh $\mathcal{M}$ with 10,475 vertices. In the following we do not use $\boldsymbol{{\theta}}_h$ and $\boldsymbol{\phi}$ and consider only the main body parameters, with a total of $22$ body joints.
We use SMPL-X instead of SMPL~\cite{loper2015smpl} to leverage the gender-neutral annotations from AMASS~\cite{AMASS}.

\myparagraph{Conditional Diffusion Models}.
\label{sec:diffusion_model_general}
We adopt the Denoising Diffusion Probabilistic Models (DDPMs) formulation~\cite{ho2020denoising}. At the core of it are a forward diffusion process and an inverse, iterative denoising process.
Let $\boldsymbol{X}_0 \sim q(\boldsymbol{X}_0)$ be the real motion distribution.
The forward diffusion process is a Markov chain adding \emph{i.i.d.} Gaussian noise at each step $t  \in \{ 1, \dots T \}$: 
\begin{equation} \label{eq:ddpm_forward}
    q(\boldsymbol{X}_t|\boldsymbol{X}_{t-1}) = \mathcal{N}(\boldsymbol{X}_t; \sqrt{\alpha_t}\boldsymbol{X}_{t-1}, (1-\alpha_t)\mathbf{I}),
\end{equation}
where $\alpha_t\in[0,1)$ defines the variance at each step according to a pre-defined schedule and $\mathbf{I}$ is the identity matrix.
~\cite{ho2020denoising} shows that $\boldsymbol{X}_t$ can be directly sampled from $\boldsymbol{X}_0$:
\begin{equation}
\label{eq:diff_sampl}
\boldsymbol{X}_t = \sqrt{\alpha_t}\boldsymbol{X}_0 + \sqrt{1-\alpha_t} \boldsymbol{\epsilon},\text{ where } \boldsymbol{\epsilon} \sim \mathcal{N}(0, \boldsymbol{I}).
\end{equation}
Starting from Gaussian noise $\boldsymbol{X}_T$, the inverse diffusion process reconstructs $\boldsymbol{X}_0$ by iteratively denoising $\boldsymbol{X}_T$ over $T$ steps. In practice, we train a denoiser neural network $D(\cdot)$ to remove the added Gaussian
noise based on condition signal $\boldsymbol{c}$ and step $t$: $\boldsymbol{\hat{X}}_0 =  D(\boldsymbol{X}_t, t, \boldsymbol{c})$ (cf.~\cite{tevet2023human}).
Leveraging Eq.~(\ref{eq:diff_sampl}), the denoiser can be trained by sampling noise step $t$ and optimizing the simple objective~\cite{ho2020denoising}:
\begin{equation}\label{eq:loss_simple}\small
\mathcal{L}_\textrm{simple} = \mathit{E}_{\boldsymbol{X}_0\sim q(\boldsymbol{X}_0|\boldsymbol{c}), t\sim [1, T]}\left[ \|\boldsymbol{X}_0 - D(\boldsymbol{X}_t, t, \boldsymbol{c})\|_2^2 \right] .
\end{equation}
In the following, we use the subscript $t$ to denote the diffusion step and $n$ to denote the frame in the motion sequence.

Note that conditioning is crucial in our setup, where we want to exploit input observations whenever available. 
\section{Method}
\label{sec:method}

\begin{figure*}
    \vspace{-2mm}
    \centering
    \includegraphics[width=\linewidth]{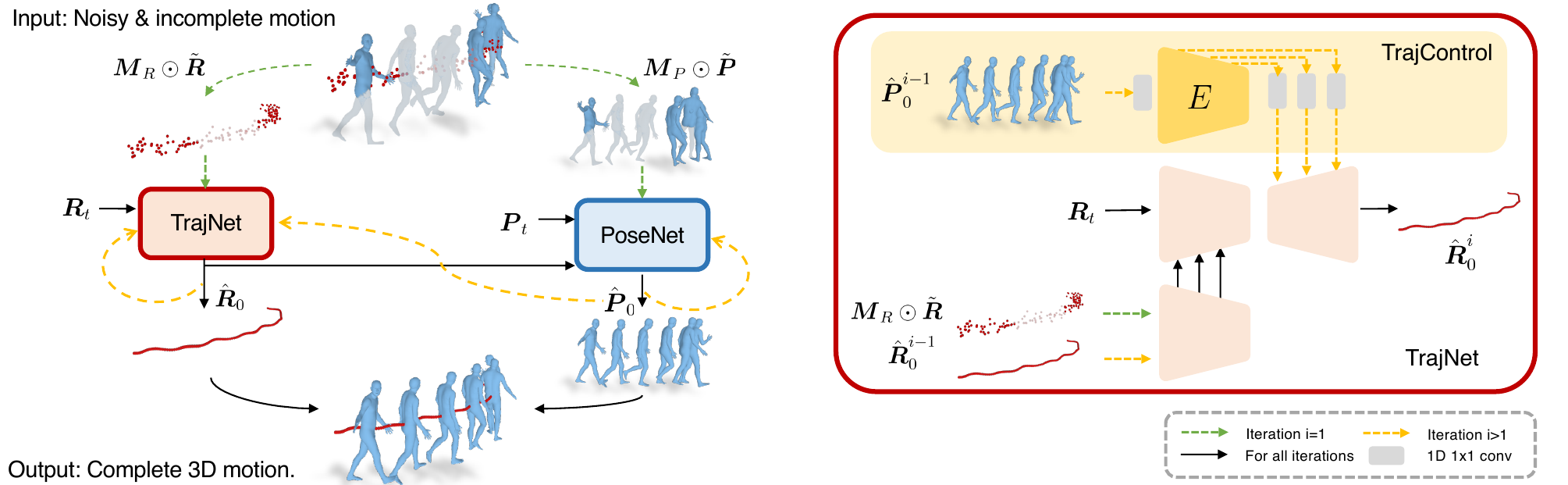}
    \vspace{-5mm}
    \caption{\textbf{Overview of our approach.} Given an initial noisy motion sequence $\tilde{\boldsymbol{X}}=(\tilde{\boldsymbol{R}}, \tilde{\boldsymbol{P}})$ and the corresponding root/body joint occlusion masks $\boldsymbol{M}_{\boldsymbol{R}}$ and $\boldsymbol{M}_{\boldsymbol{P}}$, we employ two diffusion-based models, TrajNet and PoseNet, to estimate global root trajectory $\hat{\boldsymbol{R}}_0$ and local pose $\hat{\boldsymbol{P}}_0$, separately (Sec.~\ref{sec:diffmotion}).
    We leverage an additional conditioning module, TrajControl, to fine-tune TrajNet and flexibly condition it on denoised local pose $\hat{\boldsymbol{P}}_0$, leading to improved trajectory reconstruction (Sec.~\ref{sec:trajcontrol}). 
    At inference time, TrajNet, PoseNet, and TrajControl are leveraged in an iterative inference scheme to refine local and global motion (Sec.~\ref{sec:inference}).}
    \vspace{-3mm}
    \label{fig:overview}
\end{figure*}

Our goal is to reconstruct realistic motions from monocular RGB(-D) sequences in the presence of noise and occlusions.
We use off-the-shelf, per-frame regressors~\cite{li2022cliff,feng2021collaborative, sarandi2021metrabs} and/or per-frame optimization to obtain initial SMPL-X estimates for each frame (see Sec.~\ref{sec:experiment_rgb} and Supp.~Mat. for more details). We stack these estimates into a motion sequence $\tilde{\boldsymbol{X}} \in \mathbb{R}^{N \times d}$, where $N$ is the number of frames and $d$ the dimensionality of our representation.
Such estimates are noisy, inaccurate under occlusions, and temporally inconsistent. 
Given $\tilde{\boldsymbol{X}}$, we aim to generate realistic motions $\hat{\boldsymbol{X}}_0$ in consistent global coordinates.
As reconstructing simultaneously global trajectory \emph{and} local articulated pose is challenging, we learn their dynamics with two diffusion-based models, \textit{TrajNet} and \textit{PoseNet} (Sec.~\ref{sec:diffmotion}). 
To capture correlations between the two and further refine motion plausibility, we introduce \textit{TrajControl}, a flexible conditioning module  (Sec.~\ref{sec:trajcontrol}) that we leverage with an iterative schedule at inference time (Sec.~\ref{sec:inference}). 
We describe training objectives in Sec.~\ref{sec:train}.
Fig.~\ref{fig:overview} shows an overview of our approach.

\myparagraph{Motion Representation.}
We represent an (input/output) motion sequence as $\boldsymbol{X} = (\boldsymbol{R}, \boldsymbol{P}) $, where $\boldsymbol{R} \in \mathbb{R}^{N\times d_R}$, $\boldsymbol{P} \in \mathbb{R}^{N \times d_P}$ denote root trajectory and local body features, respectively.
At each frame $n$, $\boldsymbol{x}^n = (\boldsymbol{r}^{n}, \boldsymbol{p}^{n})$ includes:
(1) global trajectory features $\boldsymbol{r}^{n}$, including root (pelvis) linear position $\boldsymbol{r}^{l} \in \mathbb{R}^{2}$, 
root angular rotation $r^a \in \mathbb{R}$, root height $r^z \in \mathbb{R}$, SMPL-X global translation $\boldsymbol{\gamma} \in \mathbb{R}^{3}$ and global orientation $\boldsymbol{\Phi} \in \mathbb{R}^{6}$;
(2) local body features $\boldsymbol{p}^{n}$, including local joint positions $\boldsymbol{J} \in \mathbb{R}^{21\times3}$, joint rotations $\boldsymbol{\theta} \in \mathbb{R}^{21\times6}$, body shape $\boldsymbol{\beta} \in \mathbb{R}^{10}$, and foot contact labels $\boldsymbol{f} \in \{0,1\}^4$:
\begin{align}
    \boldsymbol{r}^{n} &=(\boldsymbol{r}^{l}, \boldsymbol{\dot{r}}^{l}, r^a, \dot{r}^a, r^z, \boldsymbol{\gamma}, \boldsymbol{\dot{\gamma}}, \boldsymbol{\Phi}, \boldsymbol{\dot{\Phi}}), \\
    \boldsymbol{p}^{n} &=(\boldsymbol{J}, \boldsymbol{\dot{J}}, \boldsymbol{\theta}, \boldsymbol{\beta}, \boldsymbol{f}),
\end{align}
where the dot indicates the first derivative (velocity). For rotations, we adopt the 6D representation from~\cite{zhou2019continuity}. See Supp.~Mat for more details.
For $\boldsymbol{f}$, we select two joints per foot and set the corresponding contact label as 1 if the joint is in contact with the ground, 
else 0~\cite{zhang2021body, rempe2020contact}.
For each frame $n$, we define a local coordinate system such that local joint positions are relative to the current frame pelvis joint, projected on the ground~\cite{holden2016deep, tevet2023human, zhang2021learning, Guo_2022_CVPR}.
This over-parameterized representation allows explicit modelling of both 3D skeleton joints and SMPL-X meshes, such that they can benefit the one from the other.
Together with $\boldsymbol{R}$ and $\boldsymbol{P}$, we define root joint and local joint visibility masks $\boldsymbol{M}_R \in \{0,1\}^{N \times d_R}$ and $\boldsymbol{M}_P \in \{0,1\}^{N \times d_P}$, respectively (1 denotes visible, 0 otherwise). They are obtained by randomly masking joints at training time and computing joint visibility at test time (see Sec.~\ref{sec:train} and Supp.~Mat.).

\subsection{Diffusing Global and Local Motion}
\label{sec:diffmotion}

We tackle the problem of denoising and infilling global and local motion by using two networks, \emph{TrajNet} and \emph{PoseNet}. 

\myparagraph{TrajNet.} Given a noisy root trajectory $\Tilde{\boldsymbol{R}}$ and root occlusion mask $\boldsymbol{M}_R$, we train a denoiser $D_R(\cdot)$ to recover smooth and plausible global trajectory $\hat{\boldsymbol{R}}_0$:
\begin{equation}
\label{eq:trajnet}
    \hat{\boldsymbol{R}}_0 = D_R(\boldsymbol{R}_t, t, \boldsymbol{c}_R), \text{ where } \boldsymbol{c}_R = \boldsymbol{M}_R \odot \Tilde{\boldsymbol{R}}.
\end{equation}
$\boldsymbol{R}_t$ denotes root trajectory at each diffusion denoising step $t$ and $\odot$ denotes element-wise matrix multiplication.
The trajectory representation ($\boldsymbol{R}_t$, $\Tilde{\boldsymbol{R}}$, $\hat{\boldsymbol{R}}_0$) for TrajNet is parameterized as $(\boldsymbol{r}^{l}, r^a, r^z, \boldsymbol{\gamma}, \boldsymbol{\Phi})$, excluding first derivatives to avoid global drifting caused by inaccurate velocities.

\myparagraph{PoseNet.} Given denoised, infilled trajectory $\hat{\boldsymbol{R}}_0$ from TrajNet, noisy local motion $\Tilde{\boldsymbol{P}}$ and joint occlusion mask $\boldsymbol{M}_P$, we train a denoiser $D_P(\cdot)$ to recover smooth and  plausible local motion $\hat{\boldsymbol{P}}_0$:
\begin{equation}
\label{eq:posenet}
    (\hat{\boldsymbol{R}}_0, \hat{\boldsymbol{P}_0}) = D_P((\hat{\boldsymbol{R}}_0, \boldsymbol{P}_t), t, \boldsymbol{c}_P),
\end{equation}
where $\boldsymbol{P}_t$ denotes local motion at each diffusion denoising step $t$. 
The conditional signal $\boldsymbol{c}_P = (\hat{\boldsymbol{R}}_0, \boldsymbol{M}_P \odot \Tilde{\boldsymbol{P}})$ includes TrajNet's output and the corrupted initial local motion. %
To leverage the clean and complete information from TrajNet, we use the reconstructed trajectory  $\hat{\boldsymbol{R}}_0$ as both input and output for $D_P(\cdot)$, overwriting the global motion part with $\hat{\boldsymbol{R}}_0$ at each diffusion step.

\myparagraph{Architectures.}
TrajNet adopts a U-Net encoder-decoder structure built upon~\cite{rempe2023trace, janner2022planning}. An extra conditioning encoder maps the conditional trajectory signal into multi-layer features, which are concatenated with U-Net encoder features at each intermediate layer.
PoseNet employs a transformer encoder structure akin to~\cite{tevet2023human}. The conditioning signal $\boldsymbol{c}_P$ is encoded via an MLP (Multilayer Perceptron) and fed into the transformer. See Supp.~Mat. for more details.

\subsection{Controlling Global Motion Reconstruction}
\label{sec:trajcontrol}
Learning two distinct models for global and local motion does not capture the correlations between the two.
In particular, we observe that, under significant noise, TrajNet outputs might still lead to physically implausible motions with foot skating (see Sec.~\ref{sec:ablation}).
~\cite{yuan2022glamr} proposes to first infill local pose and then use that to \emph{predict} trajectory.
However, in our case we want to keep the \emph{estimated} trajectory as close as possible to input observations. Conditioning TrajNet on the initial corrupted local motion $\Tilde{\boldsymbol{P}}$ is too challenging, since such initial poses are noisy and incomplete.
Our solution is to adapt TrajNet such that it can be \emph{flexibly} conditioned on local motion estimates, when they are available.

Inspired by~\cite{zhang2023adding}, we introduce \textit{TrajControl}, an auxiliary module for fine-tuning TrajNet with additional control signal from local body pose (Fig.~\ref{fig:overview} right).
Specifically, we freeze the parameters of our pre-trained TrajNet and clone the encoder layers to a trainable duplicate, $E(\cdot)$, which serves as the \textit{TrajControl} module. 
Iteratively, we feed the output of PoseNet into $E(\cdot)$, thus improving TrajNet output based on denoised, complete local motion. In turn, the improved TrajNet output can be leveraged to further refine local body motion. We detail this iterative scheme in Sec.~{\ref{sec:inference}}.

\myparagraph{Architecture.} $E(\cdot)$ is connected to the frozen pre-trained TrajNet with zero convolution layers (1D convolution layer with kernel size 1, initialized from zero bias and zero weight), to ensure a smooth start for the fine-tuning.
During fine-tuning, we update only the weights of $E(\cdot)$. In this way, the fine-tuned TrajNet can still be conditioned on noisy trajectory only, when clean local motion is not available.

\subsection{Inference}
\label{sec:inference}
\myparagraph{Iterative inference.} We leverage TrajNet, PoseNet, and TrajControl to iteratively refine local and global motion at inference time.
Given the initial noisy/occluded motion, we first run our ``vanilla'' TrajNet and PoseNet sequentially.
TrajNet is conditioned on the initial noisy input root trajectory $\Tilde{\boldsymbol{R}}$ and root joint occlusion mask $\boldsymbol{M}_R$;
PoseNet is conditioned on the denoised trajectory $\hat{\boldsymbol{R}}_0$, noisy local motion input $\Tilde{\boldsymbol{P}}$, and body joint visibility mask $\boldsymbol{M}_P$:
\begin{align}
\label{eq:cond_trajnet_iter1}
    \boldsymbol{c}_R &= \boldsymbol{M}_R \odot \Tilde{\boldsymbol{R}}, \\
\label{eq:cond_posenet_iter1}
    \boldsymbol{c}_P &= (\hat{\boldsymbol{R}}_0^i, \boldsymbol{M}_P \odot \Tilde{\boldsymbol{P}}).
\end{align}
For any subsequent iteration $i$, TrajNet is conditioned on estimated trajectory $\hat{\boldsymbol{R}}_0^{i-1}$ and local pose $\hat{\boldsymbol{P}}_0^{i-1}$, output of PoseNet at iteration $i-1$, as the additional control signal; PoseNet is conditioned on current TrajNet output $\hat{\boldsymbol{R}}_0^i$ and local pose $\hat{\boldsymbol{P}}_0^{i-1}$ predicted at iteration $i-1$:
\begin{align}
    \label{eq:cond_trajnet_iter2}
    \boldsymbol{c}_R &= (\hat{\boldsymbol{R}}_0^{i-1}, E(t, \hat{\boldsymbol{P}}_0^{i-1})), \\
    \label{eq:cond_posenet_iter2}
    \boldsymbol{c}_P &= (\hat{\boldsymbol{R}}_0^i, \hat{\boldsymbol{P}}_0^{i-1}).
\end{align}

Algorithm~\ref{alg:inference} summarizes the approach. 
Note that at each `iteration' $i$ ($i  \in \{ 1, \dots K \}$), we sample from TrajNet and PoseNet running all $T$ diffusion denoising steps. $T$ is set to 100 for TrajNet and 1000 for PoseNet.
Empirically, we find that $K=2$ iterations are sufficient to obtain accurate results.
While one could still run this iterative inference between TrajNet and PoseNet without using TrajControl, we show in Sec.~\ref{sec:ablation} that this leads to degraded results.

\begin{algorithm}[h]
\footnotesize
\SetAlgoLined
\KwResult{Reconstructed motion ($\hat{\boldsymbol{R}}_0^K, \hat{\boldsymbol{P}}_0^K$)}
\textbf{Init}: Noisy motion ($\Tilde{\boldsymbol{R}}, \Tilde{\boldsymbol{P}}$), root occlusion mask $\boldsymbol{M}_R$, body joint occlusion mask $\boldsymbol{M}_P$, TrajNet $D_R(\cdot)$, PoseNet $D_P(\cdot)$, TrajControl $E(\cdot)$\;
 \For{$i=1:K$}{
 compute $\hat{\boldsymbol{R}}_0^i, \hat{\boldsymbol{P}}_0^i$ by Eq.~(\ref{eq:trajnet})~(\ref{eq:posenet})~(\ref{eq:cond_trajnet_iter1})~(\ref{eq:cond_posenet_iter1}) if $i=1$\;
 compute $\hat{\boldsymbol{R}}_0^i, \hat{\boldsymbol{P}}_0^i$ by Eq.~(\ref{eq:trajnet})~(\ref{eq:posenet})~(\ref{eq:cond_trajnet_iter2})~(\ref{eq:cond_posenet_iter2}) if $i>1$\;
 }
\caption{Iterative inference with TrajControl.}
\label{alg:inference}
\end{algorithm}

\myparagraph{Score-guided sampling.}
Besides embedding condition signals in the decoder architecture, diffusion models enable also test-time conditioning via classifier-based guidance, 
which has been already leveraged for image generation~\cite{sohl2015deep, song2021scorebased, dhariwal2021diffusion}, trajectory prediction~\cite{rempe2023trace}, and human mesh recovery~\cite{zhang2023probabilistic}.
In a similar spirit, we enhance physical plausibility and accuracy of reconstructed motions by guiding PoseNet sampling with two scores, penalizing foot skating and enforcing 2D joint reprojection consistency:
\begin{align}
\label{eq:score-guide-skate}
    \mathcal{J}_{\textrm{skate}}(\hat{\boldsymbol{P}}_0) &=  \|\hat{\boldsymbol{f}}_0 \dot{J}_{\textrm{3D}}^{\textrm{foot}}(\hat{\boldsymbol{R}}_0, \hat{\boldsymbol{P}}_0)\|^2, \\
\label{eq:score-guide-2d}
    \mathcal{J}_{\textrm{2D}}(\hat{\boldsymbol{P}}_0) &= \|c_{\textrm{conf}}(\Pi_\mathcal{K}(J_{\textrm{3D}}(\hat{\boldsymbol{R}}_0, \hat{\boldsymbol{P}}_0)) - J_\textrm{det})\|^2,
\end{align}
where $J_{\textrm{3D}}$ and $\dot{J}_{\textrm{3D}}$ denote body 3D joint positions and velocities obtained via forward kinematics, respectively. 
$J_\textrm{det}$ and $c_{\textrm{conf}}$ denote 2D body keypoints and their confidence scores, obtained by running OpenPose~\cite{cao2017realtime} on input images; $\Pi_\mathcal{K}$ is the 3D-to-2D projection to image space with camera intrinsics $\mathcal{K}$. The superscript `foot' identifies foot joints. $\hat{\boldsymbol{f}}_0$ denotes the foot contact labels predicted by PoseNet, so that $\mathcal{J}_{\textrm{skate}}$ penalizes foot joint velocity when the joint is predicted as touching the ground~\cite{zhang2021learning, rempe2021humor}. 
The gradients $\nabla \mathcal{J}_{(\cdot)}(\hat{\boldsymbol{P}}_0)$ effectively guide the diffusion sampling to further alleviate foot skating and better align reconstructed motion to image observations (if available). 
At each sampling step $t$, PoseNet predicts $\hat{\boldsymbol{P}}_0$ by Eq.~(\ref{eq:posenet}), which is then noised back to $\boldsymbol{P}_{t-1}$ by sampling from the Gaussian distribution:
\begin{equation}\small
   \boldsymbol{P}_{t-1} \sim \mathcal{N}(\mu_t(\boldsymbol{P}_t, \hat{\boldsymbol{P}}_0) +  (\lambda_{\textrm{skate}} \nabla \mathcal{J}_{\textrm{skate}} + \lambda_{\textrm{2D}} \nabla \mathcal{J}_{\textrm{2D}}) \Sigma_t, \Sigma_t),
\end{equation}
with $\mu_t$ as a linear combination of $\boldsymbol{P}_t$ and $\hat{\boldsymbol{P}}_0$.
The guidance is modulated by $\Sigma_t$, the variance of a pre-scheduled Gaussian distribution~\cite{ho2020denoising}, and by the scaling weights $\lambda_{\textrm{skate}}$ and $\lambda_{\textrm{2D}}$.

\subsection{Training}
\label{sec:train}
We train our diffusion denoisers $D_R(\cdot)$ and $D_P(\cdot)$ using $\mathcal{L}_{\text{simple}}$ in Eq.~(\ref{eq:loss_simple}), plus losses enforcing consistency with the ground truth in terms of 3D joint position ($\mathcal{L}_{J_{\textrm{3D}}}$) and 3D joint velocity ($\mathcal{L}_{\textrm{vel}}$), and penalizing foot skating ($\mathcal{L}_{\textrm{skate}}$):
\begin{align}
    \mathcal{L}_{J_{\textrm{3D}}} &= \|J_{\textrm{3D}}(\boldsymbol{R}_0, \boldsymbol{P}_0) - J_{\textrm{3D}}(\textcolor{blue}{\boldsymbol{R}}, \hat{\boldsymbol{P}}_0)\|^2, \\
    \mathcal{L}_{\textrm{vel}} &= \|\dot{J}_{\textrm{3D}}(\boldsymbol{R}_0, \boldsymbol{P}_0) - \dot{J}_{\textrm{3D}}(\textcolor{blue}{\boldsymbol{R}}, \hat{\boldsymbol{P}}_0)\|^2, \\
    \mathcal{L}_{\textrm{skate}} &= \|\boldsymbol{f}_0 \dot{J}_{\textrm{3D}}^{\textrm{foot}}(\boldsymbol{R}_0, \hat{\boldsymbol{P}}_0)\|^2,
\end{align}
where $(\boldsymbol{R}_0, \boldsymbol{P}_0)$ is the ground-truth motion; $\color{blue} \boldsymbol{R} $ refers to ground-truth root trajectory $\boldsymbol{R}_0$ for PoseNet, and predicted root trajectory $\hat{\boldsymbol{R}}_0$ for TrajNet.
$\boldsymbol{f}_0$ denotes ground-truth foot contact labels, and $\dot{J}_{\textrm{3D}}^{\textrm{foot}}$ denotes the predicted foot joint velocities. 
The overall loss is defined as:
\begin{equation}
\label{eq:training_objective_posenet}
    \mathcal{L} = 
    \mathcal{L}_{\textrm{simple}} + \lambda_{J_{3D}} \mathcal{L}_{J_{3D}} + \lambda_{\textrm{vel}} \mathcal{L}_{\textrm{vel}} + \lambda_{\textrm{skate}} \mathcal{L}_{\textrm{skate}},
\end{equation}

PoseNet and TrajNet are trained separately on the AMASS dataset~\cite{AMASS}, with each sequence trimmed into short clips of $N=144$ frames. For both training the ``vanilla'' TrajNet and fine-tuning TrajNet with TrajControl, we exclude $\mathcal{L}_{\textrm{skate}}$, and only compute $\mathcal{L}_{J_{3D}}$ and $\mathcal{L}_{\textrm{vel}}$ for the root joint. We utilize ground-truth local pose $\boldsymbol{P}_0$ as the control input of TrajControl to fine-tune TrajNet.

During training, we synthesize noisy and partially occluded motion $\boldsymbol{\Tilde{X}}$ by corrupting ground-truth motion $\boldsymbol{X}_0$: we add Gaussian noise to SMPL-X parameters, obtaining noisy joint positions by forward kinematics, and mask out subsets of joints.
We employ a curriculum training scheme by gradually increasing noise levels and occlusion rates as training progresses. See Supp.~Mat. for more details.

\section{Experiments}
\label{sec:experiment}
\subsection{Datasets}
\label{sec:datasets}
\myparagraph{AMASS}~\cite{AMASS} is a large-scale motion capture dataset collecting high-quality 3D human pose and shape annotations. We use the official SMPL-X neutral body annotations for training and evaluation. We downsample each sequence to 30fps.
As in~\cite{zhang2021learning}, we use TCD\_handMocap, TotalCapture, and SFU for testing and the remaining subsets for training.

\myparagraph{PROX}~\cite{PROX:2019} collects monocular RGB-D videos of people interacting with various 3D indoor scenes. Since the dataset does not provide ground-truth annotations, we use a subset of sequences to evaluate physical plausibility as in~\cite{rempe2021humor,phasemp}. 

\myparagraph{EgoBody}~\cite{Zhang:ECCV:2022} collects sequences of people interacting with each other in various 3D indoor environments, capturing multi-modal input streams with both head-mounted (first-person) and external (third-person) cameras. The dataset provides ground-truth SMPL/SMPL-X annotations.
We manually select a set of third-person RGB sequences (around 24k frames) exhibiting severe human-scene occlusions, and use them for evaluation.

\subsection{Evaluation Metrics}
\myparagraph{Accuracy.}
We adopt the Mean Per-Joint Position Error in $mm$ to evaluate predicted body joint accuracy in the pelvis-aligned coordinate system (\textit{MPJPE}) and in the \textit{global} coordinate system (\textit{GMPJPE}). We report numbers for full-body (\textit{all}), visible 
(\textit{vis}) and occluded (\textit{occ}) body joints separately, considering the 22 SMPL-X main body joints.
Furthermore, we measure foot-ground contact binary classification accuracy (\textit{Contact}) for the 4 foot joints as in ~\cite{rempe2021humor}.

\myparagraph{Physical Plausibility.}
We report additional metrics to assess motion and scene-interaction plausibility. 
When ground-truth motion is available (AMASS and EgoBody), we report the acceleration error (\textit{Accel}) as the difference in acceleration between predicted and ground-truth 3D joints~\cite{kocabas2020vibe}; for PROX, we report the norm of mean per-joint acceleration ($\|$\textit{Accel}$\|$). Both metrics are in $m/s^2$.
Foot skating ratio (\textit{Skating}) measures how often feet slide on the floor. Following~\cite{zhang2021learning}, we define sliding as happening when the velocity of all foot joints exceeds 10cm/s, toe joints' height above the ground is lower than 10cm, and ankle joints' height is lower than 15cm. 
The mean ground penetration distance (\textit{Dist})~\cite{rempe2021humor, phasemp} measures to what extent the toe joints penetrate into the ground, measured in $mm$.

\subsection{Motion from 3D Observations}
\myparagraph{Experimental setup.} 
To evaluate \mname{}'s robustness to noisy and occluded data, we run two sets of experiments on AMASS: (1) motion denoising + infilling (\textbf{Occ-L.}), and (2) motion denoising + in-betweening (\textbf{Occ-10\%}).
Given a SMPL-X motion sequence from our AMASS test set, in (1) we mask out all lower body joint parameters (both positions and rotations), simulating scenarios commonly occurring when people move in a 3D scene; in (2), we mask out an entire subsequence of frames ($10\%$ of the original sequence), thus requiring the model to generate the in-between motion.
In both setups, we add Gaussian noise to SMPL-X pose and translation parameters and use the resulting noisy and occluded 3D motion data as input for our model.
We consider different, increasing noise levels: noise level $k$ corresponds to Gaussian noise with standard deviation of $(k^{\circ}, k^{\circ}, k\ cm)$ for $(\boldsymbol{\Phi}, \boldsymbol{\theta}, \boldsymbol{\gamma})$.
Note that the noise, defined on SMPL-X parameters, will accumulate along the kinematic tree.

\myparagraph{Baselines.} 
We compare \mname{} with VPoser-t, HuMoR~\cite{rempe2021humor}, and an adapted 
version of MDM~\cite{tevet2023human} (`MDM++').
As in~\cite{rempe2021humor,phasemp}, VPoser-t is an optimization-based method using VPoser~\cite{pavlakos2019expressive} and 3D joint smoothing~\cite{huang2017towards}. It serves as the initialization stage for HuMoR.
We cannot directly apply MDM/PriorMDM~\cite{shafir2023priorMDM} to our setup, since they require noise-free visible joints for infilling and do not support explicit conditioning on noisy observations -- resulting in both pose and trajectory drifting.
We therefore train an adapted and improved version, which allows conditioning on the initial corrupted motion, using the same data augmentation as \mname{} (see Supp.~Mat. for details).

\begin{figure}
    \centering
    \includegraphics[width=\linewidth]{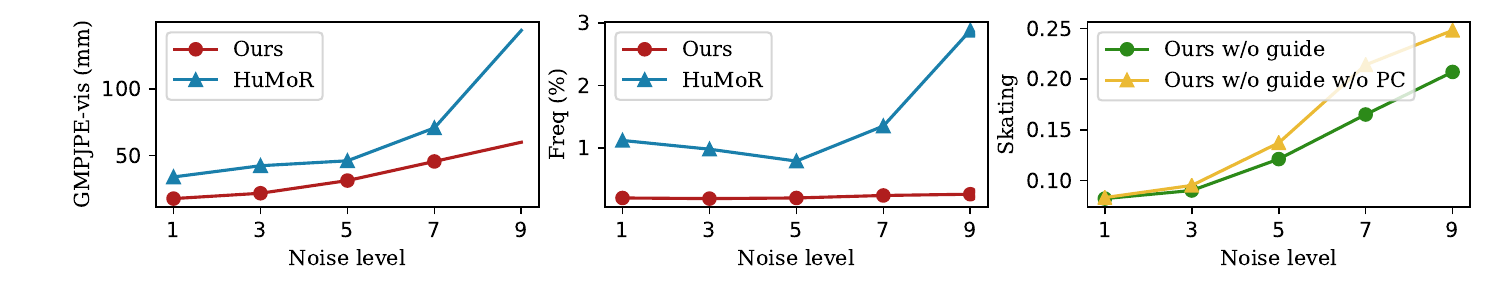}
    \caption{\textbf{Model performance wrt different input noise levels} for the \textbf{Occ-L.} setup on the AMASS test set.}
    \label{fig:exp-noise-plot}
\end{figure}

\begin{figure}
    \centering
    \includegraphics[width=\linewidth]{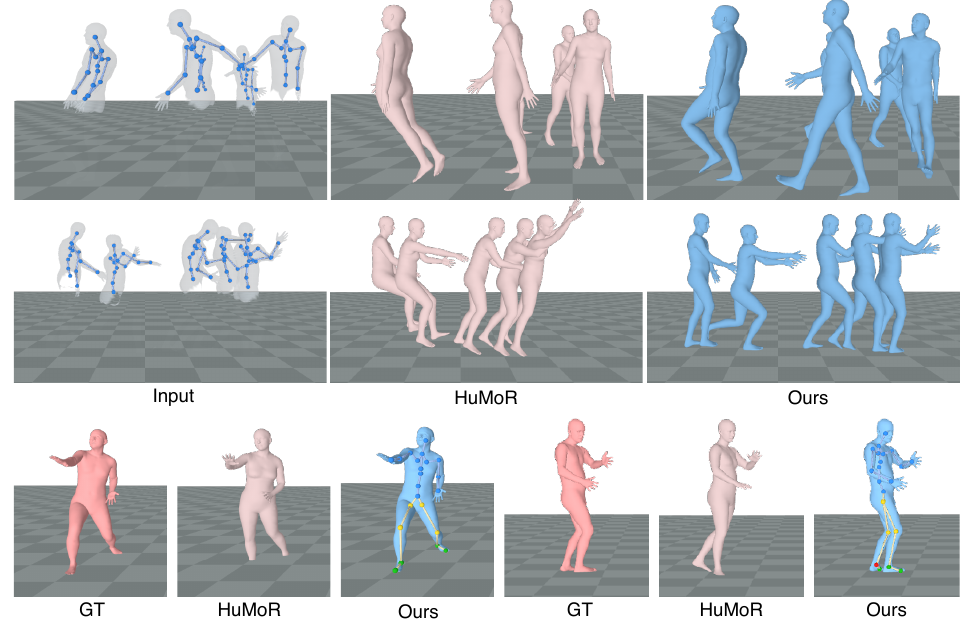}
    \caption{\textbf{Qualitative results on AMASS.}
    Given noisy input with occluded lower body, we reconstruct more accurate and realistic motions (row 1-2), with fewer foot-ground penetrations (row 3) than the baseline method.}
    \label{fig:qualitative-examples-amass}
    \vspace{-3mm}
\end{figure}

\myparagraph{Results.} 
Tab.~\ref{tab:results-amass} reports results on the AMASS test set.
Our approach demonstrates significantly superior performance, in both accuracy and physical plausibility. 
Specifically, when compared to HuMoR, we achieve $\geq$ \textbf{48\%} reduction in GMPJPE for occluded body parts in both \textbf{Occ-L.} and \textbf{Occ-10\%} setups.
The reduced acceleration errors suggest \mname{} can recover more realistic motion dynamics. This facilitates also accurate foot contact label predictions, leading to $\geq$ \textbf{44\%} improvement in foot skating over HuMoR and fewer foot-ground inter-penetrations (Fig.~\ref{fig:qualitative-examples-amass}, row 3).
MDM++ performs similarly to us in foot-ground collisions, but reconstruction accuracy and other plausibility metrics are noticeably inferior in comparison.
In scenarios with larger noise (\eg, level 7), baselines struggle to recover plausible lower body motions, often leading to legs floating in the air (see Fig.~\ref{fig:qualitative-examples-amass}, row 1-2). This is particularly evident for VPoser-t, which therefore exhibits a low skating ratio, as the skating score only considers frames with foot-ground contact.
Fig.~\ref{fig:exp-noise-plot} (left, middle) compares robustness to noise of our method and HuMoR. Increasing input noise levels corresponds to a substantial decline in performance for HuMoR, while \mname{} shows more robustness. Note that our method is only trained with a noise level of 2.

\begin{table}[ptb]
\centering
\footnotesize
\setlength{\tabcolsep}{0.5mm}
\scalebox{0.95}{
\begin{tabular}{@{}c|c|lccccccc@{}}
\toprule[1pt]
\multirow{2}{*}{Input} & \multirow{2}{*}{Noise} & \multirow{2}{*}{Method} & \multicolumn{3}{c}{GMPJPE$\downarrow$}  & \multirow{2}{*}{Accel$\downarrow$}  & \multirow{2}{*}{Cont$\uparrow$} & \multirow{2}{*}{Skat$\downarrow$}   & \multirow{2}{*}{Dist$\downarrow$}  \\
 & & & -\textit{vis} & -\textit{occ} & -\textit{all} & & & & \\

\midrule
\multirow{9}{*}{\textbf{Occ-L.}} & \multirow{3}{*}{3} & VPoser-t & 33.0 & 242.6 & 109.2 & 5.1 & - & 0.219  & 25.91 \\
 & & HuMor~\cite{rempe2021humor} & 42.4 & 167.9 & 88.0 & 3.3 & 0.68 & 0.230 & 2.59 \\
  & & MDM++ & 36.2 & 71.9 & 49.2 & 3.0 & 0.94 & 0.102 & \textbf{0.67} \\
 & & Ours & \textbf{21.8} & \textbf{57.4} & \textbf{34.8} & \textbf{2.3} & \textbf{0.95} & \textbf{0.078}  & \underline{0.69} \\
 
\cmidrule{2-10}
 & \multirow{3}{*}{5} & VPoser-t & 43.0 & 243.1 & 115.7 & 7.2 & - & 0.179 & 22.5 \\
  & & HuMor & 46.1 & 163.9 & 88.9 & 4.3 & 0.60 & 0.257 & 1.81 \\
    & & MDM++ & 40.9 & 75.4 & 53.4 & 4.4 & 0.93 & 0.126 & \underline{0.70} \\
  & & Ours & \textbf{31.3} & \textbf{66.1} & \textbf{44.0} & \textbf{3.0} & \textbf{0.94} & \textbf{0.105}  & \textbf{0.69} \\ 
  
\cmidrule{2-10}
 & \multirow{3}{*}{7} & VPoser-t & 55.1 & 247.6 & 125.1 & 9.4 & - & \textbf{0.116}  & 18.93 \\
  & & HuMor & 70.7 & 186.2 & 112.7 & 5.9 & 0.52 & 0.269 & 2.56 \\
  & & MDM++ & 53.5 & 100.0 & 77.5 & 9.3 & 0.74 & 0.287 & \textbf{0.70} \\
  & & Ours & \textbf{45.6} & \textbf{88.9} & \textbf{61.3} & \textbf{4.1} & \textbf{0.87} & \underline{0.150}  & \underline{0.76} \\ 

\midrule
\multirow{3}{*}{\textbf{Occ-10\%}} & \multirow{3}*{3} & VPoser-t & 58.9 & 136.4 & 66.4 & 3.4 & - & 0.379  & 3.12 \\
 & & HuMor~\cite{rempe2021humor} & 50.0 & 109.0 & 55.7 & 2.6 & 0.88 & 0.192  & 0.66 \\
 & & Ours & \textbf{26.3} & \textbf{56.3} & \textbf{29.2} & \textbf{2.3} & \textbf{0.96} & \textbf{0.085}  & \textbf{0.62} \\

\bottomrule
\end{tabular}
}
\caption{\textbf{Evaluation on AMASS.}
The best / second best results are in \textbf{boldface}, and \underline{underlined}, respectively. `Cont' denotes Contact, and `Skat' denotes Skating.}
\label{tab:results-amass}
\end{table}

\subsection{Motion from RGB(-D) Videos}
\label{sec:experiment_rgb}
We compare the performance of \mname{} against baselines on motion reconstruction from RGB/RGB-D videos on PROX, and from RGB videos on EgoBody.

\myparagraph{Initialization.}
On PROX, we obtain initial noisy motion estimates $\Tilde{\boldsymbol{X}}$ by running off-the-shelf per-frame 3D pose and shape regressors~\cite{li2022cliff,feng2021collaborative, sarandi2021metrabs}, returning per-frame SMPL-X parameters. 
We directly feed their output to our networks in RGB scenarios. For RGB-D sequences, we roughly align the regressor estimates to depth data via optimization minimizing joint errors. 
On EgoBody, we follow HuMoR and use VPoser-t for initialization. This allows us to perform a fair quantitative comparison against baselines -- factoring out the impact of the initialization strategy. 
Please refer to the Supp.~Mat. for more implementation details.

\myparagraph{Baselines.}
We compare our method against (1) a per-frame human mesh regressor, CLIFF~\cite{li2022cliff} (RGB only) and (2) four optimization-based methods leveraging motion priors: VPoser-t~\cite{pavlakos2019expressive,rempe2021humor}, HuMoR~\cite{rempe2021humor}, LEMO~\cite{zhang2021learning} (RGB-D only) and PhaseMP~\cite{phasemp} (RGB only)
\footnote{We compare with PhaseMP in the PROX-RGB setup using the results kindly provided by PhaseMP authors.}. 
Note that methods of type (2) are currently the ones reporting the best results for robust monocular motion reconstruction.
For reference, we also include as a baseline our initialization stage (`Ours-init').
%

\begin{table}[ptb]
\centering
\footnotesize
\scalebox{0.86}{
\begin{tabular}{@{}lccc|ccc@{}}

\toprule[1pt]
& \multicolumn{3}{c|}{RGB-D} & \multicolumn{3}{c}{RGB} \\
\cmidrule{2-7}
 Method & Skating$\downarrow$ & $\|$Accel$\|$$\downarrow$ & Dist$\downarrow$ & Skating$\downarrow$ & $\|$Accel$\|$$\downarrow$ & Dist$\downarrow$ \\
\midrule

CLIFF~\cite{li2022cliff} & - & - & - & 0.707 & 49.6 & 61.80 \\
VPoser-t & 0.286 & 3.4 & 48.75 & 0.219 & 3.2 & 50.14 \\
LEMO~\cite{zhang2021learning} & 0.176 & \textbf{1.8}  & \underline{34.22} & - & - & - \\
HuMoR~\cite{rempe2021humor} & \underline{0.117} & \underline{1.9} & 54.76 & \underline{0.139} & 2.3 & \underline{35.41} \\
PhaseMP~\cite{phasemp} & - & - & - & 0.180 & \textbf{1.8} & 46.96 \\
Ours-init & 0.565 & 24.4 & 28.70 & 0.758 & 43.7 & 73.83 \\
Ours & \textbf{0.038} & \textbf{1.8} & \textbf{3.36} & \textbf{0.116} & \underline{2.2} & \textbf{9.73} \\
\bottomrule
\end{tabular}
}
\caption{\textbf{Evaluation on PROX.}
The best / second best results are in \textbf{boldface}, and \underline{underlined}, respectively.}
\label{tab:results-prox}
\vspace{-4mm}
\end{table} 

\begin{figure}[ptb]
    \centering
    \includegraphics[width=\linewidth]{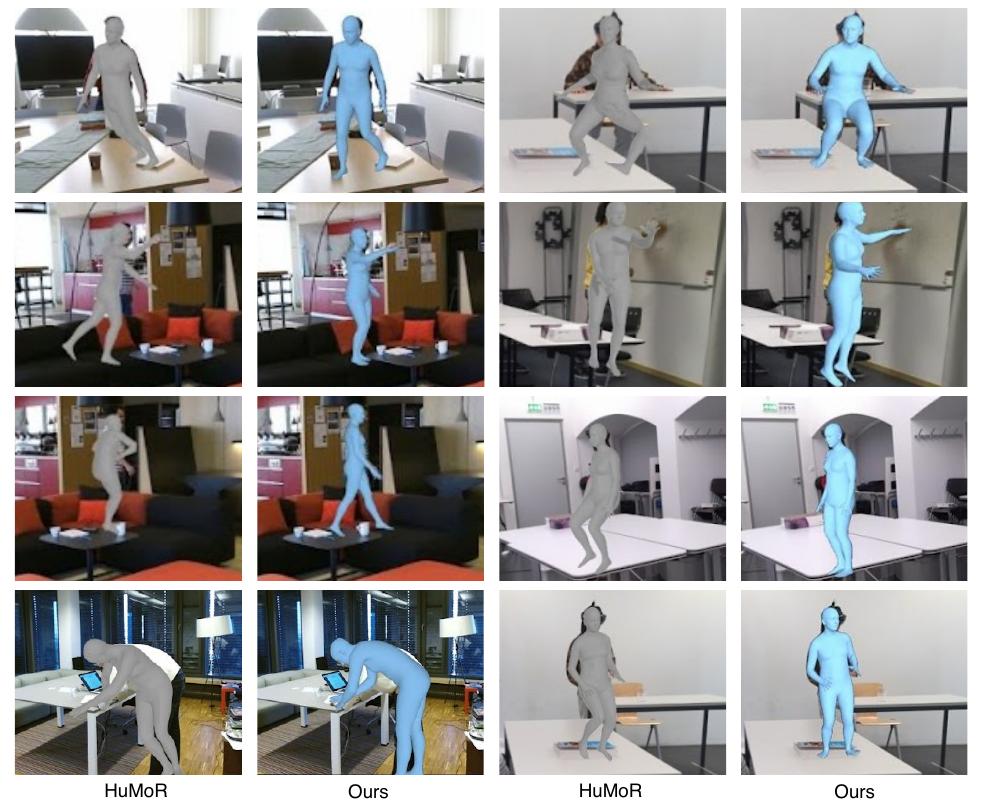}
    \caption{\textbf{Qualitative results on PROX} (RGB-D input, left) and \textbf{EgoBody} (RGB input, right).}
    \vspace{-3mm}
    \label{fig:qualitative-examples-rgb}
\end{figure}

\myparagraph{Results.} 
Tab.~\ref{tab:results-prox} reports physical plausibility results obtained on PROX.
CLIFF and Ours-init (RGB/RGB-D) are per-frame methods, producing noticeable motion jitter and foot skating. VPoser-t simply enforces 3D joint smoothness and  struggles to recover realistic motion dynamics. 
LEMO tackles noise and occlusions separately, generalizing less well to such complex scenarios. 
HuMoR and PhaseMP model motion transitions between frames but are less effective in capturing longer-range temporal correlations -- producing implausible results under heavy occlusions.
In contrast, \mname{} reconstructs smooth motions with improved foot-ground interactions (Skating and Dist), and realistic motions for occluded body parts, as shown in Fig.~\ref{fig:qualitative-examples-rgb}.
Notably, our method starts from a much more challenging initialization (Ours-init) compared to HuMoR and PhaseMP (VPoser-t), with more severe jitter and foot skating, as can be observed by comparing rows 2 and 6 in Tab.~\ref{tab:results-prox}.

Factoring out the impact of the initialization stage, Tab.~\ref{tab:results-egobody} presents quantitative results on EgoBody. We start from the same initialization as HuMoR (VPoser-t) and consistently outperform the baselines across all metrics. 
Qualitative results are shown in Fig.~\ref{fig:qualitative-examples-rgb} (right).
This indicates that \mname{} can generate more plausible motions, even in the highly occluded, challenging scenarios of this dataset. 

\noindent\textbf{Runtime}. Our approach exhibits significantly reduced runtime compared to HuMoR, being \textbf{30 times faster} factoring out the initialization stage (see details in Supp.~Mat.).

\begin{table}[tb]
\centering
\footnotesize
\scalebox{0.94}{
\begin{tabular}{@{}lcccccc@{}}
\toprule[1pt]
\multirow{2}{*}{Method} & \multirow{2}{*}{GMPJPE$\downarrow$} & \multicolumn{2}{c}{MPJPE$\downarrow$}  & \multirow{2}{*}{Accel$\downarrow$}  & \multirow{2}{*}{Skating$\downarrow$}  & \multirow{2}{*}{Dist$\downarrow$}  \\
 & & -\textit{vis} & -\textit{occ}  & & & \\

\midrule
VPoser-t & 344.8 &  \underline{63.8} &  \underline{126.2} & 3.8 &  \underline{0.143} &  \underline{13.34} \\
HuMoR~\cite{rempe2021humor} &  \underline{340.3} & 74.5 & 164.6 &  \underline{3.5} & 0.147 & 17.44  \\
Ours & \textbf{314.7} & \textbf{60.0} & \textbf{122.9} & \textbf{1.6} & \textbf{0.010} & \textbf{0.96} \\

\bottomrule
\end{tabular}
}
\caption{\textbf{Evaluation on EgoBody (RGB).}
The best / second best results are in \textbf{boldface}, and \underline{underlined}, respectively.}
\vspace{-0.7cm}
\label{tab:results-egobody}
\end{table}

\subsection{Ablation Study}
\label{sec:ablation}
\begin{figure}
    \centering
    \includegraphics[width=\linewidth]{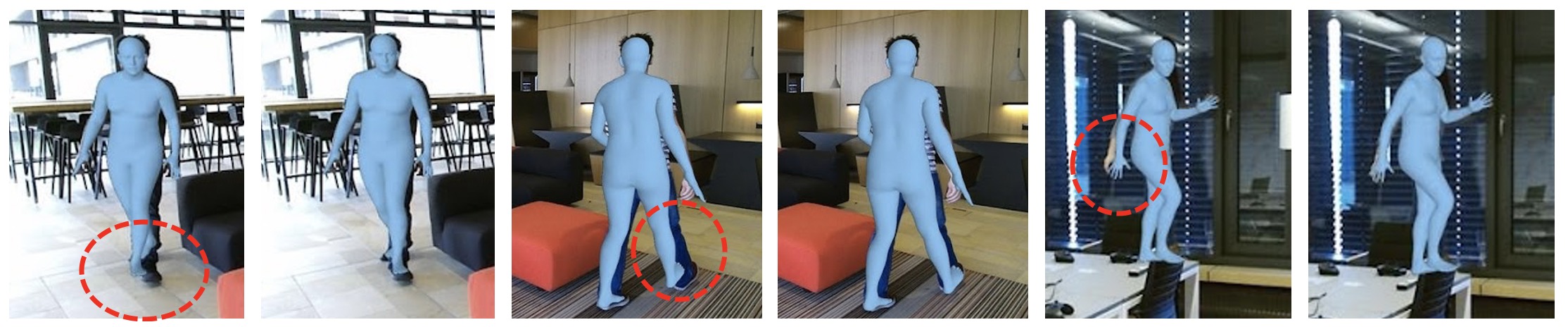}
    \caption{\textbf{Ablation for test-time guidance $ \mathcal{J}_{\textrm{2D}}$}. For each example, left/right denote without and with $\mathcal{J}_{\textrm{2D}}$ (Eq.~(\ref{eq:score-guide-2d})), respectively.}
    \vspace{-6mm}
    \label{fig:ablation-guidance-2d}
\end{figure}

We perform ablation studies on AMASS (Fig.~\ref{fig:exp-noise-plot} right, with respect to different noise levels) and PROX (Tab.~\ref{tab:ablation-prox}). 
Our iterative inference scheme leveraging TrajControl effectively alleviates foot skating by closing the gap between PoseNet and TrajNet, particularly in the presence of large noise (see Fig.~\ref{fig:exp-noise-plot} right, and `w/o TC' versus `Ours' in Tab.~\ref{tab:ablation-prox}).
Iterating between TrajNet and PoseNet twice as described in Sec.~{\ref{sec:inference}} without TrajControl (`w/o TC itr=2' in Tab.~\ref{tab:ablation-prox}) improves motion plausibility to some extent but is still sub-optimal. 
Test-time score guidance further improves result-observation alignment (see Fig.~\ref{fig:ablation-guidance-2d}) and alleviates foot skating in all setups. As expected, test-time guidance slightly impacts motion smoothness -- an aspect compensated by the iterative inference scheme.

\begin{table}[tb]
\centering
\footnotesize
\scalebox{0.85}{
\begin{tabular}{@{}lccc|ccc@{}}

\toprule[1pt]
& \multicolumn{3}{c|}{RGB-D} & \multicolumn{3}{c}{RGB} \\
\cmidrule{2-7}
 Method & Skating$\downarrow$ & $\|$Accel$\|$$\downarrow$ & Dist$\downarrow$ & Skating$\downarrow$ & $\|$Accel$\|$$\downarrow$ & Dist$\downarrow$ \\
\midrule

Ours & \textbf{0.038} & \underline{1.8} & \textbf{3.36} & \textbf{0.116} & \textbf{2.2} & \textbf{9.73} \\
 w/o TC itr=2  & \underline{0.046} & \underline{1.8}  & 4.22 & \underline{0.146} & \underline{2.3} & 10.99\\
 w/o TC  & 0.056 & 2.1 & 4.62 & 0.165 & 2.7  & 11.51 \\
 w/o $\mathcal{J}$ w/o TC & 0.072 & \textbf{1.7} & \underline{3.42} & 0.213 & \textbf{2.2}  & 10.20 \\

\bottomrule
\end{tabular}
}
\caption{\textbf{Ablation study on PROX.}
The best / second best results are in \textbf{boldface}, and \underline{underlined}, respectively. $\mathcal{J}$ denotes the test-time guidance in Eq.~(\ref{eq:score-guide-skate})(\ref{eq:score-guide-2d}), and `TC' denotes TrajControl. `itr=2' denotes two iterative iterations as in Sec.~{\ref{sec:inference}}.}
\vspace{-2mm}
\label{tab:ablation-prox}
\end{table} 

\vspace{-2mm}
\section{Conclusion}
\label{sec:conclusion}

We proposed \mname{}, an approach for robust human motion reconstruction.
Differently from previous work relying on test-time optimization~\cite{rempe2021humor,phasemp}, the approach learns how to reconstruct motion from data using diffusion models. The approach decouples the problem of recovering global and local motion by learning two models and conditioning them on available image evidence; a flexible control module captures correlations between global and local dynamics, leveraged by an iterative inference scheme to refine motion plausibility. Experiments on three publicly available datasets show that the approach can reconstruct more realistic and accurate motions than state-of-the-art baselines, especially in challenging scenarios exhibiting noise and occlusions.

\myparagraph{Limitations and Future Work.} \mname{} currently does not work online at real-time framerates. In the future, we plan to evaluate accuracy-efficiency tradeoffs using different architectures (\eg,~\cite{du2023agrol}). 
Moreover, adapting \mname{} to further incorporate 3D environment conditions to model human-scene interactions is an exciting avenue for future research.
Finally, while here we focused on full-body reconstruction, future work should extend \mname{} to model also facial expressions and articulated hand poses over time.

{
\myparagraph{Acknowledgements.} 
Siyu Tang is partially funded by the SNSF project grant 200021 204840.
We sincerely thank Korrawe Karunratanakul, Sebastian Starke, Marko Mihajlovic, Tony Tung, Yuting Ye, Artsiom Sanakoyeu, and Yuhua Chen for insightful discussions.
}

{
\small
\bibliographystyle{ieeenat_fullname}
\bibliography{main}
}

\clearpage
\begingroup
\onecolumn 

\appendix

\begin{center}
\Large{\bf RoHM: Robust Human Motion Reconstruction via Diffusion \\
**Supplementary Material**\\
}
\vspace{0.2cm}
\vspace{0.6cm}
\end{center}

\setcounter{page}{1}
\setcounter{table}{0}
\setcounter{figure}{0}
\renewcommand{\thetable}{S\arabic{table}}
\renewcommand\thefigure{S\arabic{figure}}


\section{Architecture Details}
The detailed architecture of our model is illustrated in Fig.~\ref{fig:appendix-architecture}. 

\myparagraph{TrajNet} adopts a U-Net structure built upon~\cite{rempe2023trace, janner2022planning}, with a series of 1D temporal convolutional blocks (`ConvBlock') to downsample and upsample the input root trajectory $\boldsymbol{R}_t$ at diffusion denoising step $t$, and predict the clean trajectory $\hat{\boldsymbol{R}}_0$. The U-Net encoder and decoder are connected via skip connections. 
At each inference iteration $i$ (Sec.~4.3 in the main paper), an extra conditioning encoder, structured similarly to the U-Net encoder, encodes the trajectory signal ($\hat{\boldsymbol{R}}_0^{i-1}$ for inference iteration $i>1$, \textcolor{yellow}{yellow arrow}, and $\boldsymbol{M}_R \odot \Tilde{\boldsymbol{R}}$ for $i=1$, \textcolor{forestGreen}{green arrow}) into multi-layer features.
These features are concatenated with the intermediate U-Net encoder features at each convolutional block. These two parts constitute the ``vanilla'' TrajNet.

\myparagraph{TrajControl}  models pose-trajectory correlations and further refines root trajectory (Sec.~4.2 in the main paper), based on the denoised and infilled local body pose $\hat{\boldsymbol{P}}_0^{i-1}$ from the previous inference iteration $i-1$.
Namely, upon completing the training of the vanilla TrajNet, the U-Net encoder, along with its weights, is duplicated to serve as the TrajControl encoder, to encode pose information. The intermediate pose features are added to the U-Net decoder via zero convolution layers (1x1 convolution with weights and bias initialized from zero). The TrajControl module is fine-tuned while keeping other TrajNet components frozen. This ensures that the vanilla TrajNet can process input even when only a corrupted trajectory is provided.

\myparagraph{PoseNet} builds on the transformer encoder architecture from~\cite{tevet2023human}. At each diffusion denoising step $t$ during inference iteration $i$, the input local pose $\boldsymbol{P}_t$ is concatenated with the estimated trajectory from TrajNet, $\hat{\boldsymbol{R}}_0^{i}$, and then fed into the transformer encoder. 
Regarding the conditioning signal, body pose (corresponding to the corrupted pose $\boldsymbol{M}_P \odot \Tilde{\boldsymbol{P}}$ for iteration $i=1$, \textcolor{forestGreen}{green arrow}, and the estimated body pose $\hat{\boldsymbol{P}}_0^{i-1}$ for iteration $i>1$, \textcolor{yellow}{yellow arrow}) is combined with the estimated trajectory $\hat{\boldsymbol{R}}_0^{i}$ and processed through a linear embedding layer. This conditioning feature, along with the embedding of the diffusion step $t$, serves as the input to the transformer encoder.

\begin{figure}[h]
    \centering
    \includegraphics[width=0.8\linewidth]{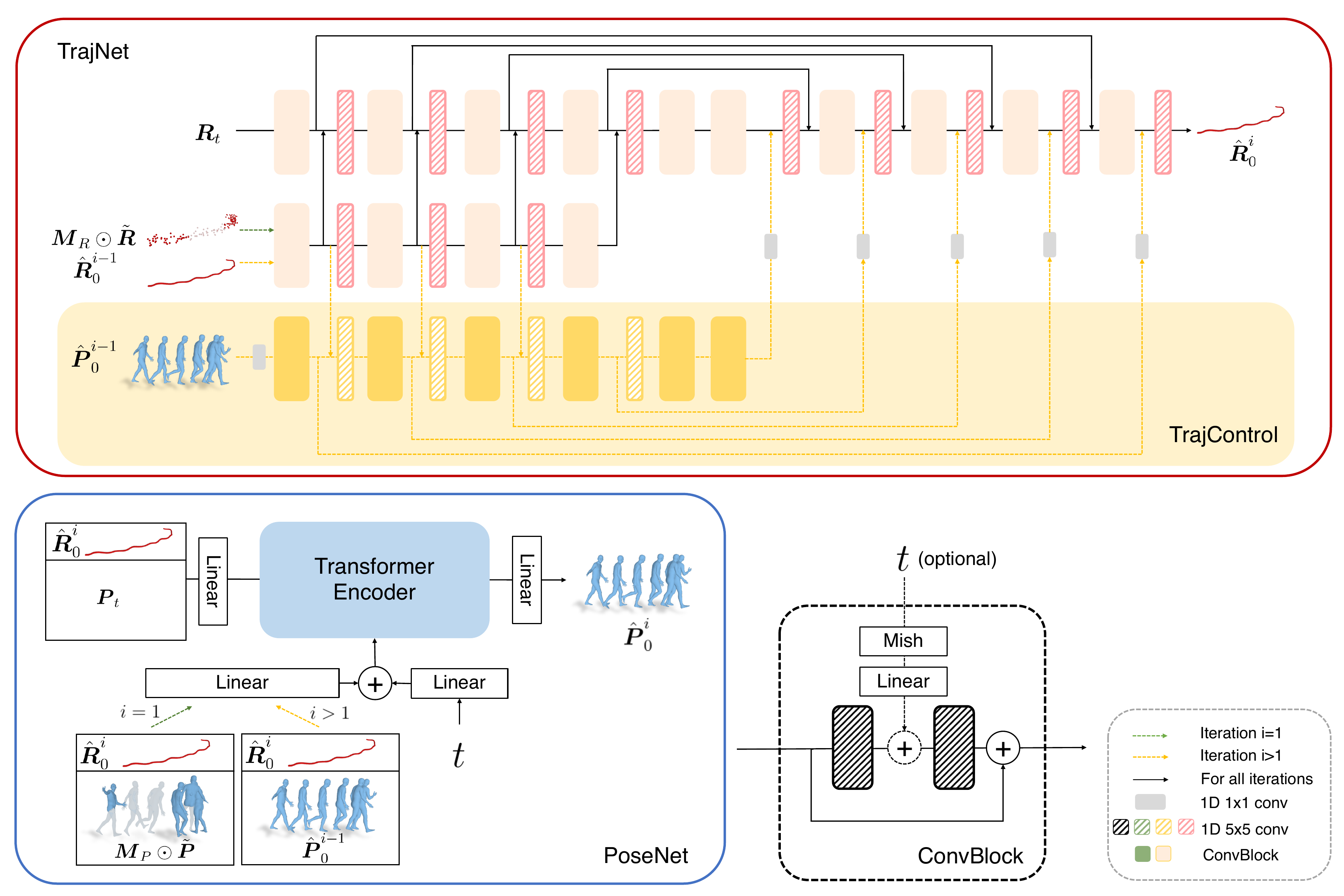}
    \caption{\textbf{Model architecture for TrajNet, TrajControl, and PoseNet.} 
    }
    \label{fig:appendix-architecture}
     \vspace{-0.2cm}
\end{figure}

\section{Implementation Details}

\subsection{Training Details}
\label{sec:appendix-augmentation}
\myparagraph{Data augmentation.}
During training, we apply Gaussian noise and synthetic occlusion masks to ground-truth motion sequences from AMASS~\cite{AMASS}, to simulate noisy and occluded input motion $\Tilde{\boldsymbol{X}}$.
We add Gaussian noise with a noise level $k$ to the ground-truth SMPL-X parameters, with zero mean and standard deviation of $(k^{\circ}, k^{\circ}, k\ cm, 0.01k)$ for $(\boldsymbol{\Phi}, \boldsymbol{\theta}, \boldsymbol{\gamma}, \boldsymbol{\beta})$; we then obtain noisy 3D joint positions via forward kinematics. 
During the initial training phases, the model is trained on easier tasks, with lower noise levels and smaller occlusion ratios for $\boldsymbol{\Tilde{X}}$. As the training progresses, we gradually expose the model to harder cases, with higher noise levels and heavier occlusions, as detailed below.

\myparagraph{Training schemes.}
TrajNet undergoes training in four stages. In the initial two stages, the model is trained with noise level $k=1$ and $k=2$, respectively, without occlusions. In the third stage, the noise level is raised to $k=3$, with and 10\% of the frames entirely masked out. Upon completion of these stages, the training for the vanilla TrajNet is concluded. In the last stage, TrajControl is fine-tuned to incorporate additional control from body pose, with a noise level of $k=2$ and no occlusion masks.
PoseNet follows a two-stage training process. In the first stage, the model is trained with a noise level of $k=1$. To synthesize occlusion masks, we randomly mask out 1-6 joints in the initial 500 epochs. Afterwards, a mixed occlusion scheme is applied: with 0.5 probability, occlusion masks from PROX pseudo ground truth are used; with 0.3 probability, all lower body parts are masked out; with 0.2 probability, all upper body parts are masked out; with 0.1 probability, the full body is masked out in 30\% of the frames. In the second stage, we continue the mixed occlusion scheme and increase the noise level to $k=2$.

\myparagraph{Noise assumptions.}
We experiment with various training noise scales, and Tab.~\ref{tab:appendix-noise-distribution} confirms the synthetic noise distribution on AMASS given by the selected scale aligns with real-world input noise (given by VPoser-t initialization on EgoBody). Synthetic noise applied to SMPL-X parameters accumulates along the kinematic tree, which mirrors real-world scenarios: running VPoser-t on EgoBody yields larger MPJPEs for wrists (0.23m) than shoulders (0.12m).

\begin{table}[h]
\centering
\scalebox{0.94}{
\begin{tabular}{ccccc}
\toprule[1pt]
MPJPE   & 0-0.2m & 0.2-0.4m & 0.4-0.6m & $>$0.6m \\
\midrule
AMASS (synthetic noise) & 78.1\% & 18.9\% & 2.7\%  &0.3\%  \\
EgoBody (real-world noise) & 82.0\%  &13.7\%   &3.5\%   &0.8\%  \\
\bottomrule
\end{tabular}
}
\caption{\textbf{The synthetic noise distribution aligns with real-world noise:} percentage of joints whose GMPJPE (between noisy / GT joints) falls into each range. 
}
\vspace{-0.3cm}
\label{tab:appendix-noise-distribution}
\end{table}

\myparagraph{Training weights.} 
For both PoseNet and TrajNet, weights $\lambda_{\textrm{3D}}$ and $\lambda_{\textrm{vel}}$ are set to 100 and 1000, respectively. $\lambda_{\textrm{skate}}$ is set to 0 during the first training stage, and 0.1 during the second training stage in PoseNet.

\myparagraph{PoseNet Training strategy.}
PoseNet is trained with the GT trajectory instead of TrajNet output, as the small error between them (pelvis GMPJPE 12.5 mm) ensures minimal impact. Training PoseNet with TrajNet output requires separate trainings with vanilla and fine-tuned TrajNet, thus not optimal for efficiency.

\subsection{Motion Representation}
In the proposed motion representation, the root linear position $\boldsymbol{r}^{l}$ and height $r^z$ denote the pelvis $xy$ coordinates projected on the ground, and pelvis $z$ coordinate, respectively. There is a shape-dependent shift between $(\boldsymbol{r}^{l}, r^z)$ and the SMPL-X body translation $\boldsymbol{\gamma}$.
The root angular rotation $r^a$ refers to the body rotation around z-axis. This is similar to SMPL-X global orientation $\boldsymbol{\Phi}$, but not identical. 
Existing works~\cite{tevet2023human, karunratanakul2023guided} typically adopt only the joint-based representation, and resort to time-consuming post-processing optimization to obtain the expressive SMPL body meshes.
Our over-parameterized representation enables learning of motion dynamics in both the joint space and SMPL-X parameter space, such that the model can directly predict the SMPL-X bodies without any post-processing step.
Although the ultimate goal is to output the SMPL-X body mesh, the joint-based representation part is necessary, as the joint space is more straightforward and explicit for the motion learning, which in turn benefits the learning for SMPL-X parameters.

\subsection{Motion Initialization}
For experiments on PROX~\cite{PROX:2019}, we utilize the per-frame body regressor CLIFF~\cite{li2022cliff} to estimate per-frame body poses from RGB input as initialization. In contrast to most existing human mesh regressors, which take as input only an image cropped around the human body, CLIFF incorporates information from the cropped bounding box (scale and location with respect to the original image) and the original image focal length. This approach yields improved predictions for global orientation, particularly beneficial when the body is positioned at the boundary of the original input image. However, it is worth noting that CLIFF is trained on the SMPL body model. 
Consequently, we complementarily employ a SMPL-X based human mesh regressor, PIXIE~\cite{feng2021collaborative}, to estimate also SMPL-X body shape parameters $\boldsymbol{\beta}$. We then combine the pose from CLIFF and the shape from PIXIE. Additionally, to enhance global translation estimation, we leverage a skeleton-based 3D human pose regressor, MeTRAbs~\cite{sarandi2021metrabs}, which provides a better prediction for the absolute global position.
We combine the global orientation and body pose obtained from CLIFF, the body shape derived from PIXIE, and the global translation estimated by MeTRAbs and use them as our per-frame initialization $\Tilde{\boldsymbol{X}}$ for motion estimation from RGB videos on PROX.

For RGB-D sequences on PROX, we additionally perform a per-frame optimization step to incorporate depth observations. More precisely, for each frame, we optimize the SMPL-X body parameters $(\boldsymbol{\Phi}, \boldsymbol{\theta}, \boldsymbol{\gamma}, \boldsymbol{\beta})$ by minimizing the following objective function:
\begin{equation}
    \mathcal{L} = \lambda_{\textrm{2D}} \mathcal{L}_{\textrm{2D}} + \lambda_{\textrm{depth}} \mathcal{L}_{\textrm{depth}} + \lambda_{\textrm{pose}} \mathcal{L}_{\textrm{pose}} + \lambda_{\textrm{shape}} \mathcal{L}_{\textrm{shape}}.
\end{equation}
$\mathcal{L}_{\textrm{2D}}$ penalizes the 2D joint distances between the optimized 2D SMPL-X body joints projected onto the RGB image, and detections from OpenPose~\cite{cao2017realtime}. $\mathcal{L}_{\textrm{depth}}$ penalizes the 3D Chamfer distance between the human point cloud obtained from the
depth frame and SMPL-X surface points visible from the
camera as in~\cite{PROX:2019, zhang2021learning}. $\mathcal{L}_{\textrm{pose}}$ and $\mathcal{L}_{\textrm{shape}}$ denote priors that regularize SMPL-X body pose and shape. $\lambda$s denote the corresponding weights. This approach is akin to VPoser-t but excludes the 3D joint smoothness term, working per-frame. 

On EgoBody~\cite{zhang2022egobody}, to conduct a quantitative comparison with the baselines while factoring out the influence of various initialization strategies, we employ VPoser-t for initialization as in  HuMoR~\cite{rempe2021humor}. 
Regarding the input OpenPose 2D detections for our method and baseline methods, instead of raw detections, we use a manually post-processed version provided by EgoBody, where the detections for most occluded joints are masked out.

It is worth highlighting that our approach can be combined with various initialization strategies (both optimization- and regression-based), 
ensuring flexibility for different applications and inputs.

\subsection{Inference Details}
\myparagraph{Occlusion masks for reconstruction from RGB(-D) videos.}
To obtain joint occlusion masks for inference on PROX and EgoBody, given the initialized 3D body, we identify a body joint as occluded if it fulfills two conditions: (1) the confidence score of the corresponding 2D joint detection is below 0.2; and (2) the depth of the joint is greater than the depth of the scene vertex which is projected on the same 2D pixel in the image plane as the body joint, from the camera view. The depth of the joint is determined by rendering the 3D body mesh obtained from initialization from the camera view.

\myparagraph{Score-guided sampling.}
In Eq.~(14) in the main paper, we set $\lambda_{\textrm{2D}}$ to 3e5.
$\lambda_{\textrm{skate}}$ is set to 1e5 for experiments on PROX and EgoBody, and to 3e6 for experiments on AMASS. The score-guided sampling is enabled for the last 100 denoising steps for PoseNet. Furthermore, as the modulation variance $\Sigma_t$ diminishes towards the end of the diffusion denoising steps, we skip the last 20 denoising steps for PoseNet for experiments on PROX and EgoBody; this ensures stronger gradient guidance for 2D alignment with image observations.

\myparagraph{Runtime.} 
To assess the runtime difference between our method and HuMoR~\cite{rempe2021humor}, we omit the initialization stage (as the runtime depends on the setup and the choice of the initialization method), and focus solely on the inference/test-optimization stage for both methods. 
For RGB-D input, employing an NVIDIA A100 GPU with a batch size of 10, and with a sequence length of 144 frames, our method completes the inference in 59 seconds, while HuMoR requires 30 minutes for the entire test-time optimization. We use the default configurations of the official HuMoR code. Note that the run time we report here differs from the ones in HuMoR paper due to the different hardware and setups (batch size and sequence length).
Besides, we also report the inference time under the same setup as above for CLIFF/VPoser-t as 0.5sec, and 2.5min, respectively.

\section{Baseline MDM++ Details}
For motion infilling and in-between tasks, at each denoising step, MDM~\cite{tevet2023human} and PriorMDM~\cite{shafir2023priorMDM} replace denoised joints with visible input joints, when they are available.
This assumes clean motion for visible body parts as input, and therefore cannot handle noisy scenarios like the ones we consider. Moreover, we observe that the relative trajectory representation in~\cite{tevet2023human, shafir2023priorMDM}, which only considers trajectory velocities, results in severe global trajectory drifting and deviation from the input, due to accumulated errors in the estimated trajectory velocities. To address these limitations and enable denoising together infilling and in-between tasks, we adapt the original MDM formulation to obtain MDM++, as explained below.

MDM++ shares a similar design with PoseNet (Fig.~\ref{fig:appendix-architecture}), but with two key distinctions. 
Firstly, MDM++ takes the initial noisy and incomplete motion $(\boldsymbol{M}_R \odot \Tilde{\boldsymbol{R}}, \boldsymbol{M}_P \odot \Tilde{\boldsymbol{P}})$ as the condition, and concurrently predicts both root trajectory $\hat{\boldsymbol{R}}_0$ and local body pose $\hat{\boldsymbol{P}}_0$. This means that, differently from ~\cite{tevet2023human, shafir2023priorMDM}, 
MDM++ explicitly conditions on noisy motion by taking noisy trajectory and local pose as input -- thus enabling motion denoising at inference time. We train MDM++ with the same augmentation scheme as \mname{}, see Sec.~\ref{sec:appendix-augmentation}.
Secondly, MDM++ adopts the same motion representation as our method, as detailed in Sec.~4 of the main paper, incorporating both the absolute and relative representations for the root trajectory. This design choice significantly mitigates trajectory drifting issues.

However, addressing both denoising and infilling tasks in two different spaces (root trajectory and local pose) within one single model remains very challenging. Indeed, MDM++ still exhibits degraded reconstruction accuracy and motion plausibility compared to \mname{}, as shown in Tab.~1 in the main paper.

\section{More Results}
\myparagraph{Results on unoccluded data.}
Here we report RoHM reconstruction accuracy (MPJPE) for unoccluded frames (frames with all body joints visible) on the EgoBody subset (the same subset as used in Sec.~5 in the main paper) as 60.1mm, while HuMoR yields an MPJPE of 73.1mm for unoccluded data.
This indicates that, apart from cases with both noise and occlusions, our reconstruction also outperforms the baseline in accuracy specifically for scenarios with noise only but without occlusions.

\myparagraph{Cross-view results.} Fig.~\ref{fig:appendix-cross-view} illustrates that the cross-view results are also plausible.

\begin{figure}[h]
    \centering
    \includegraphics[width=\linewidth]{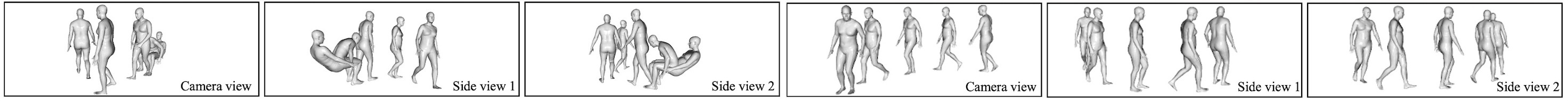}
    \caption{\textbf{Cross view results for two sample sequences.}}
    \label{fig:appendix-cross-view}
    \vspace{-0.2cm}
\end{figure}

\myparagraph{Thresholds for foot skating.}
To investigate the impact of different thresholds in the Skating metric, we evaluate the foot skating ratio on EgoBody with different velocity thresholds, and RoHM consistently outperforms baselines: VPoser-t / HuMoR / Ours: 0.19/0.17/\textbf{0.04} with threshold 15cm/s, 0.22/0.23/\textbf{0.08} with 10cm/s, 0.28/0.39/\textbf{0.23} with 5cm/s.

\myparagraph{Non-determinism.}
Despite the stochastic nature of generative models, our reconstruction accuracy is not impacted by the non-determinism due to constraints posed by visible joints and the temporal span (144 frames). For the setup in Tab.~1 (Occ-L., noise level 3) in the main paper, different random seeds yield similar results: GMPJPE\textcolor{blue}{\textit{-occ}}/\textcolor{pink}{-\textit{vis}} are \textcolor{blue}{within 57.3$\sim$57.7mm} / \textcolor{pink}{constantly 21.8mm}.

 \vspace{-0.1cm}
\section{Limitations and Failure Cases}
We show example failure cases in Fig.~\ref{fig:appendix-failure-cases}. As it is common for learning-based approaches, our method can struggle to generalize to out-of-distribution test cases -- such as shapes and poses that are rarely seen in the training data. For instance, the first two columns of Fig.~\ref{fig:appendix-failure-cases} show subjects that are relatively tall, and the last two columns show rare poses. 

Another limitation lies in the model's dependence on both the 3D scene mesh and 2D joint detections to determine if a joint is occluded. This reliance becomes problematic when the 3D scene mesh is unavailable or when 2D joint detections are unreliable. A potential solution could involve learning an occlusion classifier based on the initial 3D body pose and image inputs to identify joint occlusions. We consider this avenue a promising direction for future exploration.

 \vspace{-0.2cm}
\begin{figure}[h]
    \centering
    \includegraphics[width=0.6\linewidth]{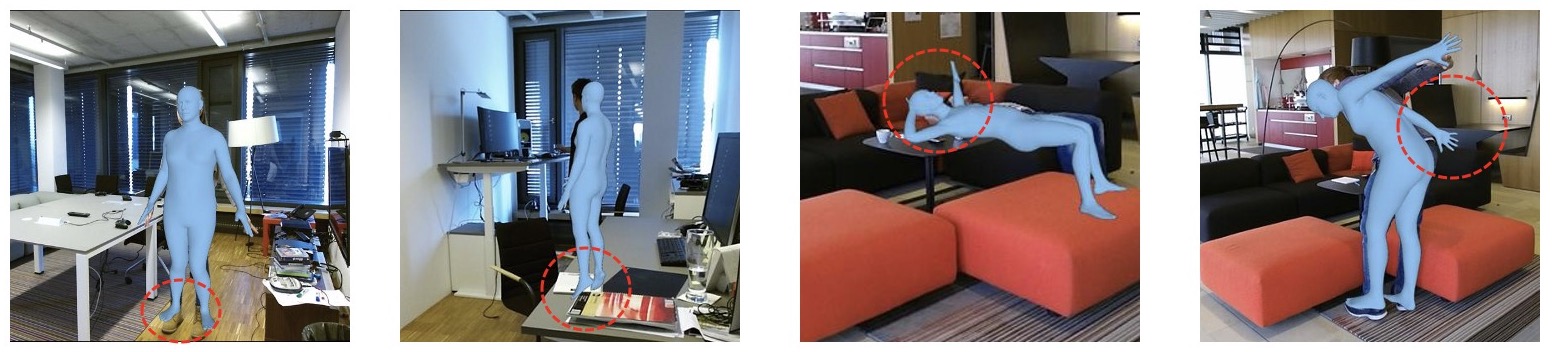}
    \vspace{-0.3cm}
    \caption{\textbf{Failure cases} with inaccurate estimations for out-of-distribution shapes (column 1, 2) and poses (column 3, 4).}
    \label{fig:appendix-failure-cases}
    \vspace{-0.2cm}
\end{figure}


\end{document}